\documentclass[]{bytedance}
\usepackage[toc,page,header]{appendix}

\usepackage{minitoc}
\usepackage{amsfonts}
\usepackage{amssymb}
\usepackage{tabularx}
\usepackage{listings}
\usepackage{xcolor}
\usepackage{cancel}
\usepackage{wrapfig}

\usepackage{tabulary,multirow,xspace}
\usepackage{fixmath,mathtools,nicefrac,mmstyle}
\usepackage{subcaption}
\usepackage{amssymb}   

\newcommand{\iconEdit}{$\spadesuit$}      
\newcommand{\iconReason}{$\diamondsuit$}  

\newcommand{\sE}{\textsuperscript{\,\iconEdit}}
\newcommand{\sR}{\textsuperscript{\,\iconReason}}
\newcommand{\sER}{\textsuperscript{\,\iconEdit\iconReason}}
\usepackage{caption}
\usepackage{wrapfig} 
\usepackage[misc]{ifsym} 
\usepackage{colortbl}

\usepackage{wrapfig}
\usepackage{multicol}
\usepackage[most]{tcolorbox}
\usepackage{pifont}
\usepackage{arydshln}   
\usepackage{array}
\newcolumntype{C}{>{\centering\arraybackslash}p{1.2cm}}  

\definecolor{codegreen}{rgb}{0,0.6,0}
\definecolor{codegray}{rgb}{0.5,0.5,0.5}
\definecolor{codepurple}{rgb}{0.58,0,0.82}
\definecolor{backcolour}{rgb}{0.95,0.95,0.92}
\definecolor{boxblue}{RGB}{57,89,163}
\definecolor{boxbluebg}{RGB}{230,237,250} 
\definecolor{myblue}{RGB}{210, 225, 255}

\lstdefinestyle{mystyle}{
    backgroundcolor=\color{backcolour},   
    commentstyle=\color{codegreen},
    keywordstyle=\color{magenta},
    numberstyle=\tiny\color{codegray},
    stringstyle=\color{codepurple},
    basicstyle=\ttfamily\footnotesize,
    breakatwhitespace=false,         
    breaklines=true,                 
    captionpos=b,                    
    keepspaces=true,                 
    numbers=none,                    
    numbersep=5pt,                  
    showspaces=false,                
    showstringspaces=false,
    showtabs=false,                  
    tabsize=2
}
\definecolor{mygray1}{gray}{.95}
\definecolor{mygray2}{gray}{.9}
\definecolor{mygray3}{gray}{.95}
\usepackage{pifont}

\newlength\savewidth
\newcolumntype{x}[1]{>{\centering\arraybackslash}p{#1pt}}

\newcommand{\app}{\raise.17ex\hbox{$\scriptstyle\sim$}}

\usepackage{xcolor}
\usepackage{graphicx}
\usepackage{amssymb}
\usepackage{pifont}
\usepackage{floatrow}
\usepackage{amsmath} 
\usepackage{float}
\usepackage{wrapfig}
\usepackage{multirow}
\usepackage{tcolorbox}
\tcbuselibrary{breakable, skins, raster}
\usepackage{listings}
\usepackage{listings}
\usepackage[absolute,overlay]{textpos}

\definecolor{commentgreen}{rgb}{0.1, 0.4, 0.1}
\definecolor{keywordblue}{rgb}{0.1, 0.1, 0.7}
\definecolor{stringred}{rgb}{0.7, 0.1, 0.1}

\lstdefinestyle{mystyle}{
    commentstyle=\color{commentgreen},
    keywordstyle=\color{keywordblue},   
    stringstyle=\color{stringred},
    basicstyle=\ttfamily\scriptsize, 
    breaklines=true,
    keepspaces=true,
    showstringspaces=false,
    frame=none,                     
    language=Python, 
}

\title{From Storage to Access: Verifiable Activation of Parametric Knowledge in LLMs via \\Explicit Priming and Implicit Reasoning}

\author[1,2, \star]{Zuocheng Ying}
\author[1, \star,\dagger]{Yang Yang}
\author[1]{Yumou Wu}
\author[1]{Chuanbo Zhu}
\author[1]{Jiarui Wang}
\author[1]{\\Ziqi Wu}
\author[1]{Jingming Cai}
\author[2]{Junqing Yu}
\author[2, \dagger]{Zikai Song}

\affiliation[1]{ByteDance}
\affiliation[2]{Huazhong University of Science and Technology}
\contribution[\star]{Equal contribution}
\contribution[\dagger]{Corresponding author}

\abstract{ 
Although Large Language Models (LLMs) encode rich factual knowledge in their parameters, reliably recalling and verifying such knowledge remains a key bottleneck in factual question answering. Existing end-to-end methods entangle knowledge elicitation with reasoning, making it difficult to determine whether correct answers arise from parametric knowledge or the input context. 
To address this challenge, we propose VAKE (\textit{\textbf{V}erifiable \textbf{A}ctivation of Parametric \textbf{K}nowledg\textbf{E}}), a two-stage reinforcement-learning framework that externalizes latent parametric knowledge through explicit Priming and transfers the acquired elicitation capability to implicit Reasoning. Given a query and an insufficient retrieved subgraph, the \textbf{Priming} policy explicitly inserts bridging triples as verifiable evidence, with supervision provided by rewards derived from answers generated by a separate frozen model over the augmented subgraph. 
Building on the policy learned during Priming, the \textbf{Reasoning} stage trains the model to answer from the original input, testing whether the capability acquired through explicit knowledge elicitation transfers to implicit reasoning.
Experiments across seven benchmarks and models from 3B to 14B show that VAKE consistently outperforms standard baselines, including when transferring directly from HotpotQA to OOD datasets. 
LLM-based evaluation further shows that over 80\% of the inserted triples provide factual bridging knowledge not derivable from the retrieved context, while more than half elicit knowledge inaccessible through direct prompting. 
These results suggest that VAKE activates latent parametric knowledge rather than copying the input context or memorizing dataset-specific associations.
}

\correspondence{Yang Yang at \email{yang.yves@bytedance.com}, Zikai Song at \email{skyesong@hust.edu.cn}}

\begin{document}

\begin{textblock*}{2cm}(16cm,1.85cm)   
  \includegraphics[height=1cm]{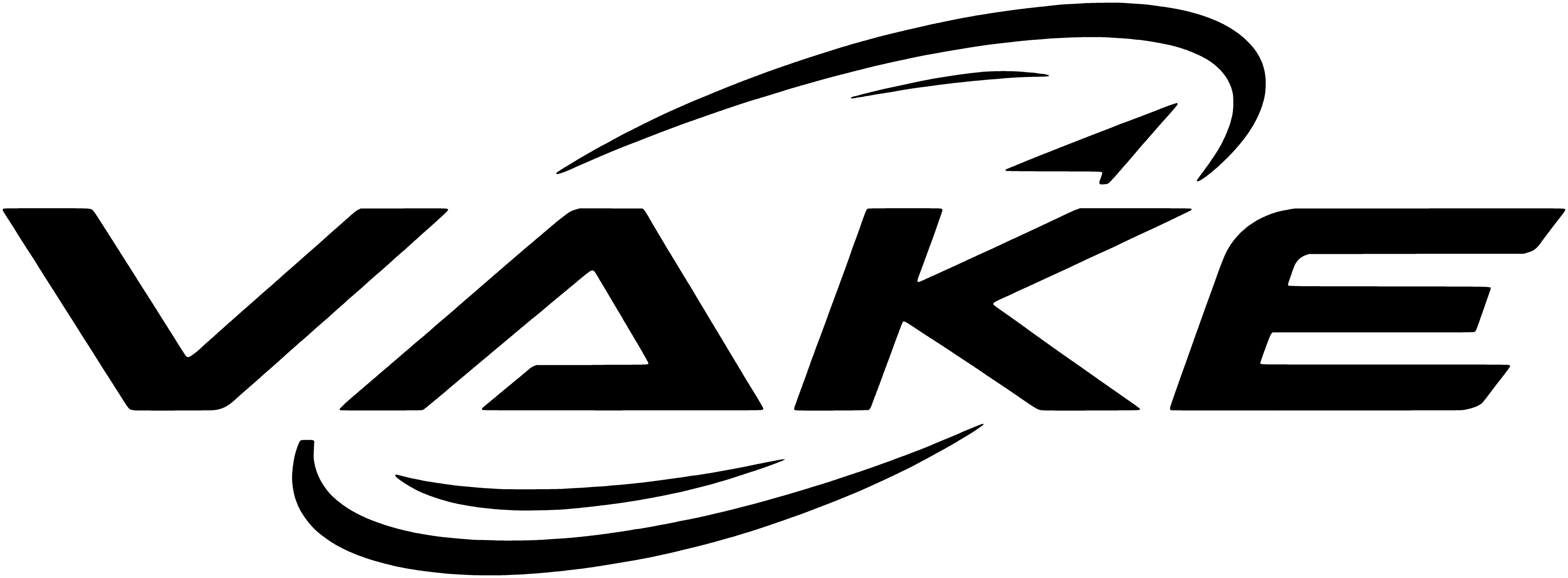}
\end{textblock*}

\maketitle

\section{Introduction}
To equip large language models (LLMs) with general world knowledge and usable reasoning ability, researchers adopt a standard two-stage training pipeline. Pre-training encodes vast factual knowledge into LLMs’ parameters, while post-training methods, such as supervised fine-tuning (SFT) and reinforcement learning (RL), adapt this parametric knowledge to task-specific behaviors and outputs \cite{deepseekai2026deepseekv4highlyefficientmilliontoken, glm5team2026glm5vibecodingagentic, yang2025qwen3technicalreport}. 
Prior literature suggests that frontier models achieve near-saturated encoding of factual knowledge from encyclopedic corpora such as Wikipedia \cite{calderon2026empty, gekhman2026thinking}. 
However, a critical bottleneck remains: models may fail to recall factual knowledge within their parameters, an issue defined as ``stored but inaccessible” \cite{zheng2024reliablellmsknowledgebases, wang-etal-2024-unveiling}.

Recall-oriented evaluations further expose this limitation: models may fail to answer queries correctly even when the target facts are fully present in the training corpus \cite{zheng2024reliablellmsknowledgebases, mousavi2026doeslossoptimizationactually}. Prior analyses suggest that internal parametric representations store richer factual information than what greedy decoding can reliably elicit, yet fact-association recall remains fragile \cite{wang-etal-2024-unveiling, gekhman2025insideout}. Scaling model size and training data can alleviate encoding deficits, but yields limited gains in accessing latent parametric knowledge \cite{calderon2026empty, mousavi2026doeslossoptimizationactually}.
\begin{figure}[t]
\centering
\includegraphics[width=0.7\columnwidth]{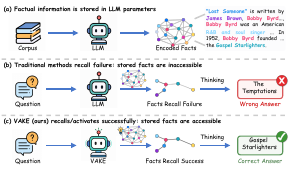}
\caption{
    VAKE activates inaccessible parametric knowledge. (a) Factual information is encoded in LLM parameters. (b) Traditional methods suffer from fact recall failure: encoded facts remain inaccessible during thinking, leading to ungrounded reasoning and a wrong answer. (c) VAKE activates the relevant factual information before thinking, enabling fact recall success and the correct answer.
}
\label{fig1}
\end{figure}

This limitation is difficult to diagnose with standard accuracy-based evaluations, which conflate recall failures with encoding failures despite requiring different interventions. Encoding failures motivate scaling or knowledge injection, whereas recall failures call for post-training methods that improve access to knowledge already encoded in model parameters. Existing methods use instruction tuning (See Figure \ref{fig1}) and reinforcement learning to improve knowledge utilization \cite{yang2026reasoningreinforcementlearningunlocks, meng2026sparse}. However, their reliance on unstructured generation entangles knowledge elicitation with answer reasoning, making the effect of activation difficult to isolate \cite{calderon2026empty, mousavi2026doeslossoptimizationactually}. Crucially, eliciting latent parametric knowledge often requires external query-relevant cues to trigger the model's fact-association recall. To address this, we draw inspiration from the “priming effect” in cognitive science: a phenomenon where exposure to specific cues activates latent memory representations, enhancing subsequent task performance. Transferring this to LLMs, we hypothesize that introducing an independent, cue-conditioned activation phase over an insufficient retrieved subgraph can awaken the model's dormant parametric knowledge.

To this end, we propose \textbf{V}erifiable \textbf{A}ctivation of Parametric \textbf{K}nowledg\textbf{E} (VAKE), a two-stage reinforcement-learning framework that separates explicit knowledge elicitation from direct answer reasoning.
VAKE consists of \textbf{Priming} and \textbf{Reasoning}. In Priming, an RL-optimized policy elicits latent parametric knowledge as relational triples and inserts them into a sparse retrieved subgraph. 
A separate, frozen answerer then evaluates this augmented graph and provides outcome-based rewards.
This insert-then-answer intervention ensures that any improvement in the answerer’s output is directly attributable to the explicitly injected triples, providing a verifiable reward signal for knowledge activation.
In Reasoning, the policy initialized from Priming is optimized via GRPO to answer directly from the original retrieved subgraph without explicitly injected triples. This effectively internalizes the acquired elicitation capability into implicit chain-of-thought (CoT) reasoning. 
Experiments show that VAKE consistently outperforms non-retrieval activation baselines, standard GRPO, and prior RL-based activation methods across model scales. Models optimized on a single multi-hop dataset transfer effectively to diverse out-of-distribution (OOD) multi-hop and single-hop datasets. Applying GRPO after Priming yields further gains, demonstrating the complementarity of knowledge activation and reasoning optimization. Therefore, VAKE can be seamlessly integrated into existing post-training pipelines as an RL-based complement and SFT-based knowledge activation.

In summary, our main contributions are as follows:
\begin{itemize}
    \item \textbf{A two-stage framework that separates knowledge activation from direct answer reasoning.} Priming explicitly elicits latent parametric knowledge as relational triples, while Reasoning transfers the learned elicitation capability to direct question answering.

    \item \textbf{An observable and attributable representation of activated knowledge.} Relational triples provide discrete and inspectable evidence for evaluating the content and effect of knowledge activation.

    \item \textbf{Compatibility with standard post-training paradigms.} Empirically validated by our ablation results, VAKE serves as an RL-based complement to standard knowledge activation and seamlessly integrates into standard post-training pipelines while retaining full compatibility with reasoning optimization.
\end{itemize}

\section{Related Work}
\label{sec:related}

\subsection{Access, Not Storage}
Factual QA is increasingly framed as an \textit{access} problem rather than a
\textit{storage} one, because a model's greedy output only lower-bounds what it
encodes. Three lines of evidence support this framing. External behavior shows
that direct prompting underestimates recall
\citep{zheng2024reliablellmsknowledgebases,wang-etal-2024-unveiling}, that
models answer inconsistently across paraphrases
\citep{jiang-etal-2020-know,elazar-etal-2021-measuring}, and that fine-tuning
on new facts can even hurt existing recall \citep{gekhman-etal-2024-fine}.
Internal probing shows that hidden states carry truthfulness signals absent
from the decoded answer
\citep{burns2023discovering,azaria-mitchell-2023-internal,NEURIPS2023_81b83900,orgad2025llms}
and that internal knowledge measurably exceeds what the model explicitly
generates \citep{gekhman2025insideout}. Controlled provenance further shows
that an injected corpus can make a fact provably parametric yet still leave the
model unable to surface it
\citep{mousavi2026doeslossoptimizationactually,ovadia2025knowledge,liu-etal-2025-structure}.
We therefore study multi-hop QA instances limited by access to encoded
knowledge, and seek an activation procedure that improves such access.

\subsection{Eliciting Activation and Its Confounds}
Existing activation methods fall into two families, and both leave the same
measurement confound in which the effect of activation cannot be cleanly
separated from reasoning or retrieval. \textbf{Inference-time} methods produce
intermediate content at query time through prompt search
\citep{shin-etal-2020-autoprompt,zhong-etal-2021-factual}, \textit{RECITE}
\citep{sun2023recitationaugmented}, \textit{Self-Ask}
\citep{press-etal-2023-measuring} and \textit{Step-Back}
\citep{zheng2024take}, all built on chain-of-thought and generated-knowledge
prompting
\citep{NEURIPS2022_9d560961,liu-etal-2022-generated,wang2023selfconsistency}.
\textbf{Training-time} methods use reinforcement learning, where a correctness reward reweights knowledge the model holds rather than new facts \citep{yang2026reasoningreinforcementlearningunlocks}, and token-level and
distributional analyses take the same view that RLVR sharpens the base
distribution rather than expanding it
\citep{meng2026sparse,yue2025does,wen2026reinforcement}.
Related RL variants either route access through a reasoning chain
\citep{gekhman2026thinking,ma2026improving,chen2025learning}
or combine access with side objectives such as factuality
\citep{ren-etal-2026-knowrl,li2025reasoning,wei2025truthrl}.
Because both families operate over free-form text and provide no controlled
context, activation can only be inferred from output changes rather than
directly attributed.

We take a route that removes this confound. We provide the model with an
incomplete retrieved subgraph that offers relevant cues without a sufficient
answer path, and keep the answerer frozen during Priming, so the inserted
bridging triples are the only free variable and their answer-enabling effect
is observable and attributable. We treat the subgraph as a medium for
query-conditioned augmentation rather than an external structure to traverse,
and learn the Priming policy with reinforcement learning.
\section{Method}
\label{sec:method}

\subsection{Problem Formulation}
\label{sec:problem-formulation}
Let $M_0$ be a frozen model whose parameters encode knowledge that may not be directly accessible for a given query. For a question $q$ with ground-truth answer $a^\star$ and an insufficient context $c$, our goal is to elicit the required knowledge from model parameters and express it as an explicit, query-conditioned increment $\mathcal{I}$. A parameterized policy $\pi_\theta$ generates this increment as
\begin{equation}
    \mathcal{I}\sim\pi_\theta(\cdot\mid q,c).
\end{equation}
We define successful activation by the following model-relative intervention:
\begin{equation}
    M_0(q,c) \neq a^\star,\qquad
    M_0(q,c\cup\mathcal{I}) = a^\star .
\end{equation}



Here, the knowledge is \textit{parametric} in source but becomes \textit{explicit} in $\mathcal{I}$. Unlike retrieval augmentation, $\mathcal{I}$ is generated by the policy rather than retrieved from an external source. Its attribution and answer-enabling effect are evaluated through the controlled intervention above.

We implement this activation via two stages: \textbf{Priming} and \textbf{Reasoning}.
Drawing on the cognitive notion of priming, where cues facilitate access to latent information, \textbf{Priming} trains the policy to elicit implicit parametric knowledge. We externalize the elicited knowledge as answer-enabling bridging triples $\mathcal{I}$, making activation observable and operationally attributable.
\textbf{Reasoning} then trains the policy to perform CoT reasoning that exploits this elicitation capability and corrects hallucinations, enabling direct question answering without explicit triple insertion.
\textcolor{red}{
}



\subsection{Task Definition}
\label{sec:task-definition}

We instantiate the parametric-knowledge activation task defined above on a triple graph $\mathcal{G}=\{(s,r,o)\}$ parsed offline from the associated documents. In the main protocol, raw documents are used only for graph construction and are not exposed during training or standard inference; the model observes only triples. A retriever returns a query-relevant subgraph $\mathcal{R}(q)\subseteq\mathcal{G}$, which serves as the context $c=\mathcal{R}(q)$. A few diagnostic experiments explicitly add raw documents to Priming, but the answerer always receives only triples.
The retriever is designed to be shallow and sparse, providing query-relevant cues that can condition the elicitation of parametric knowledge rather than a complete supporting path. The concrete retrieval configuration is described in Sec.~\ref{sec:experimental-setup}.

\begin{figure*}
\centering
\includegraphics[width=\textwidth]{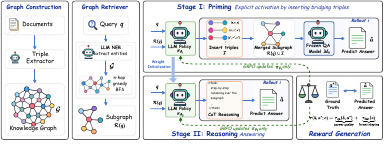} 

\caption{
Overview of our proposed VAKE.
Given a sparse query-relevant subgraph $\mathcal{R}(q)$, VAKE first performs explicit priming by learning to insert verifiable bridging triples $\mathcal{I}$ for a frozen answerer $M_0$ (Stage I).
It then transfers the learned knowledge-elicitation capability to implicit CoT reasoning through GRPO in the Reasoning stage (Stage II).
}
\label{fig3}
\end{figure*}



\subsection{Two-Stage Activation: Priming and Reasoning}

The two stages operate on the same input $(q,\mathcal{R}(q))$ and optimize the same answer-quality objective with respect to $a^\star$, but differ in how the answer is produced. In the \textbf{Priming} stage, the policy inserts explicit bridging triples as intermediate evidence for a frozen answerer. In the \textbf{Reasoning} stage, the Priming-initialized policy generates the answer directly from the original input, transferring the learned knowledge-elicitation capability to direct reasoning. The two stages are optimized sequentially. Figure~\ref{fig3} gives an overview of this two-stage activation process.



\subsubsection{stage I: Priming (insert-then-answer)}

In this stage, the policy $\pi_\theta$ acts as a subgraph-augmentation policy. Given the query and the retrieved subgraph, it externalizes bridging facts as a set of inserted triples
\begin{equation}
    \mathcal{I} \sim
    \pi_\theta(\cdot \mid q,\mathcal{R}(q)),
\end{equation}
after which a frozen answerer $M_0$ produces
\begin{equation}
    \hat{a}
    =
    M_0\bigl(q,\mathcal{R}(q)\cup\mathcal{I}\bigr).
\end{equation}

This insert-then-answer intervention makes knowledge activation observable, attributable, and rewardable. With $M_0$ and $\mathcal{R}(q)$ held fixed, $\mathcal{I}$ is the only free variable; a transition from an incorrect answer without $\mathcal{I}$ to a correct one with $\mathcal{I}$ therefore isolates the effect of the inserted knowledge. Attribution is further supported by comparing each inserted triple against $\mathcal{R}(q)$: when the decisive evidence is absent and non-derivable from the retrieved subgraph, the gain cannot be explained by copied context and is attributed to knowledge elicited from the policy's parameters. Priming thus needs no SFT-style labels for inserted triples; its reward comes only from whether the downstream frozen answerer uses them to produce $a^\star$.





\subsubsection{stage II: Reasoning}

The insert-then-answer decomposition in stage I is an instrument for making activation attributable, rather than the intended final usage mode of the model. After Priming, we return the policy to its native question-answering role. Initialized from the stage I policy,
\begin{equation}
    \theta \leftarrow \theta^{(1)},
\end{equation}
the model now answers directly,
\begin{equation}
    \hat{a} \sim \pi_\theta(\cdot \mid q, \mathcal{R}(q)),
\end{equation}
without an inserted set or a frozen answerer, and is optimized by GRPO with the same input setting and reward design as Priming, but with a stage-specific format-shaping term. This stage tests whether the explicit activation induced by Priming can be carried into chain-of-thought answering.

This second stage also tests whether gains from Priming are complementary to reasoning gains. If Priming merely reproduces the effect of standard chain-of-thought optimization, continuing with reasoning-oriented GRPO should yield limited additional improvement. Additional gains instead indicate complementarity under sequential optimization. We therefore report Priming followed by Reasoning as a core result rather than as a mere ablation, since it examines whether Priming can be incorporated before reasoning-oriented GRPO in existing post-training pipelines.

\subsection{Reinforcement Learning Objective}

Both stages in Section~\ref{sec:method} are trained by reinforcement learning and optimized with the same GRPO algorithm. They share an answer-quality reward against the gold answer $a^\star$, but differ in how $\hat{a}$ is produced and which output format is required. We keep the reward outcome-driven but shape it into a small number of interpretable components rather than a single hard $0/1$ signal. For stage $s$, the overall reward is: 
\begin{equation}
    \label{eq:reward}
    r^{(s)}(\hat{a}, a^{\star}; o)
    \;=\;
    r_{\mathrm{qa}}(\hat{a}, a^{\star})
    \;+\;
    r_{\mathrm{fmt}}^{(s)}(o).
\end{equation}
The answer-quality term $r_{\mathrm{qa}}$ is a soft blend of exact match (EM) and token-level F1 between $\hat{a}$ and $a^\star$, weighted so that exact correctness dominates while the F1 component densifies the reward landscape and mitigates the sparsity of a pure $0/1$ signal on hard multi-hop questions. The stage-specific format-shaping term $r_{\mathrm{fmt}}^{(s)}$ depends only on whether the trajectory $o$ is well formed: it grants a small bonus when the output follows the required triple or reasoning format, and imposes a penalty when it is malformed, with severe malformation overriding the quality term so that unparseable trajectories cannot be rewarded.

The two stages instantiate $\hat{a}$ differently. In \textbf{Priming}, the trainable policy emits an inserted triple set and the answer is produced by the frozen answerer,
\begin{equation}
    \hat{a} = M_0\bigl(q, \mathcal{R}(q) \cup \mathcal{I}\bigr),
\end{equation}
so the reward reaches the policy only through $\mathcal{I}$. In \textbf{Reasoning}, the policy answers directly,
\begin{equation}
    \hat{a} \sim \pi_\theta(\cdot \mid q, \mathcal{R}(q)),
\end{equation}
and the reward scores its own output.

The concrete weights, bonuses, and penalty magnitudes in Eq.~\eqref{eq:reward} are detailed in the \textit{Supplementary Material}.

\subsubsection{Group-Relative Policy Optimization}

We optimize both Priming and Reasoning with GRPO. For each question $q$, the old policy $\pi_{\theta_{\mathrm{old}}}$ samples a group of $G$ trajectories $\{o_i\}_{i=1}^{G}$. In Priming, each trajectory specifies inserted bridging triples, and the frozen answerer $M_0$ predicts $\hat{a}_i$ from the subgraph after triple insertion. In Reasoning, the trajectory directly contains the reasoning trace and final answer. Each trajectory receives the stage-specific reward $r_i=r^{(s)}(\hat{a}_i,a^\star;o_i)$.

Following GRPO, we compute the group-relative advantage by normalizing rewards within the sampled group:
\begin{equation}
\label{eq:group-advantage}
\hat{A}_i =
\frac{r_i-\operatorname{mean}(\{r_j\}_{j=1}^{G})}
{\operatorname{std}(\{r_j\}_{j=1}^{G})+\delta},
\end{equation}
where $\delta$ is a small constant for numerical stability. The same sequence-level advantage is assigned to all tokens in $o_i$.

Let $\rho_{i,t}(\theta)$ be the token-level probability ratio between
$\pi_\theta$ and $\pi_{\theta_{\mathrm{old}}}$, and define
\begin{equation}
\label{eq:grpo-clip}
c_{i,t}(\theta)=\min\!\left\{\rho_{i,t}(\theta)\hat{A}_i,\;
\operatorname{clip}_{\varepsilon}\!\left(\rho_{i,t}(\theta)\right)\hat{A}_i\right\}.
\end{equation}
The GRPO objective is
\begin{equation}
\label{eq:grpo-objective}
\mathcal{J}(\theta)
=
\mathbb{E}\!\left[
\frac{1}{G}\sum_{i=1}^{G}
\frac{1}{|o_i|}\sum_{t=1}^{|o_i|}
\left(c_{i,t}(\theta)-\beta D_{i,t}^{\mathrm{KL}}\right)
\right],
\end{equation}
where $\varepsilon$ controls the clipping range, $\beta$ weights the KL penalty,
and $D_{i,t}^{\mathrm{KL}}$ denotes the token-level KL penalty to the
stage-specific reference policy.

Gradients are propagated only through $\pi_\theta$. During Priming, $M_0$ is fixed and non-differentiable, contributing only through the reward induced by its generated answer. Thus, optimization learns to insert more useful bridging triples without adapting the answerer to compensate for weak insertions. During Reasoning, no separate answerer is used, and the policy is rewarded directly by the quality of its generated answer.
\section{Experiments}
\label{sec:exp}


\subsection{Experimental Setup}
\label{sec:experimental-setup}

\begin{table*}[t]
\centering
\small
\begin{tabular*}{\textwidth}{@{\extracolsep{\fill}} l CCCC CCCCC}
\toprule
& \multicolumn{4}{c}{\textbf{In-Distribution Multi-Hop QA}}
& \multicolumn{5}{c}{\textbf{Out-of-Distribution QA}} \\
\cmidrule(lr){2-5}\cmidrule(lr){6-10}
\textbf{Method}
& \textbf{2Wk}$^\dagger$
& \textbf{Hotp}$^\dagger$
& \textbf{MuS}$^\dagger$
& \textbf{Avg}
& \textbf{Bam}$^\dagger$
& \textbf{NQ}$^\star$
& \textbf{Triv}$^\star$
& \textbf{Pop}$^\star$
& \textbf{Avg} \\
\hline

\rowcolor{gray!10}
\multicolumn{10}{c}{\textbf{\textit{Qwen2.5-7B-Instruct}}} \\

Base
    & 27.2 & 32.7 & 13.4 & 24.4
    & 12.0 & 28.7 & 51.6 & 18.0 & 27.6 \\

Base\sR
    & 25.2 & 33.2 & 13.6 & 24.0
    & 21.6 & 28.3 & 51.2 & 17.8 & 29.7 \\

Self-Ask
    & 29.8 & 38.6 & 18.4 & 28.9
    & \textbf{41.6} & 30.6 & 55.4 & 18.0 & 36.4 \\

RECITE
    & 30.7 & 36.3 & 17.6 & 28.2
    & \underline{40.8} & \textbf{34.8} & \underline{57.4} & 19.8
    & \underline{38.2} \\

Unlock
    & 29.4 & 35.3 & 16.6 & 27.1
    & 12.8 & 28.5 & 52.0 & 18.3 & 27.9 \\

GRPO
    & 32.5 & 35.4 & 16.2 & 28.0
    & 12.8 & 28.7 & 51.8 & 18.2 & 27.9 \\

GRPO\sR
    & 33.5 & 31.9 & 15.8 & 27.1
    & 30.4 & 28.0 & 52.5 & 18.9 & 32.5 \\

\hdashline

VAKE-P\sE
    & 33.1 & 40.9 & 20.0 & 31.3
    & 32.0 & 30.8 & 54.0 & 18.8 & 33.9 \\

VAKE\sER
    & \textbf{36.2} & \underline{42.3} & \textbf{21.0} & \textbf{33.2}
    & 34.4 & 31.1 & 56.2 & \underline{20.1} & 35.5 \\

VAKE\sR
    & \underline{35.5} & \textbf{43.4} & \underline{20.4} & \underline{33.1}
    & \textbf{41.6} & \underline{33.2} & \textbf{59.5} & \textbf{21.2}
    & \textbf{38.9} \\

\hline

\rowcolor{gray!10}
\multicolumn{10}{c}{\textbf{\textit{Qwen3-8B}}} \\

Base
    & 28.2 & 36.2 & 14.8 & 26.4
    & 14.4 & 30.8 & 50.9 & 19.6 & 28.9 \\

Base\sR
    & 26.5 & 33.5 & 18.4 & 26.1
    & \underline{40.8} & 33.8 & 60.1 & 22.3 & 39.3 \\

Self-Ask
    & 34.8 & 38.5 & 19.8 & 31.0
    & 32.8 & 27.9 & 53.0 & 18.5 & 33.1 \\

RECITE
    & 33.0 & 40.1 & 19.6 & 30.9
    & 33.6 & 35.3 & 58.5 & 21.5 & 37.2 \\

Unlock
    & 31.4 & 38.5 & 21.6 & 30.5
    & 15.2 & 32.0 & 51.9 & 20.3 & 29.9 \\

GRPO
    & 34.2 & 36.8 & 20.2 & 30.4
    & 16.8 & \underline{35.7} & \underline{61.4} & \textbf{22.5}
    & 34.1 \\

GRPO\sR
    & 36.2 & 41.8 & 19.0 & 32.3
    & \underline{40.8} & 34.1 & 61.2 & \underline{22.4}
    & \underline{39.6} \\

\hdashline

VAKE-P\sE
    & 35.0 & 42.3 & \underline{22.4} & 33.2
    & 32.0 & 31.3 & 59.7 & 21.8 & 36.2 \\

VAKE\sER
    & \underline{37.1} & \underline{43.7} & \textbf{23.8} & \textbf{34.9}
    & 36.0 & 34.4 & 61.0 & 22.2 & 38.4 \\

VAKE\sR
    & \textbf{37.9} & \textbf{44.9} & 20.6 & \underline{34.5}
    & \textbf{43.2} & \textbf{37.4} & \textbf{62.3} & \textbf{22.5}
    & \textbf{41.4} \\

\bottomrule
\end{tabular*}
\caption{
    \textbf{Answer accuracy (\%) on ID and OOD QA benchmarks.}
    $^\dagger$ and $^\star$ indicate multi-hop and single-hop QA datasets, respectively. 
    The badges denote the test-time pipeline:
    \iconEdit{}~=~bridging-triple insertion and
    \iconReason{}~=~CoT reasoning.
    For each backbone group, \textbf{best} and \mbox{\underline{second-best}} results are highlighted.
}
\label{tab:main_results}
\end{table*}

\textbf{Datasets.} We evaluate on two groups. (i) \textit{In-distribution}: three multi-hop datasets, 2WikiMultihopQA \cite{ho2020constructing}, HotpotQA \cite{yang2018hotpotqa}, and MuSiQue \cite{trivedi2022musique}, used for both training and testing. (ii) \textit{Out-of-distribution (OOD)}: 
four held-out datasets used only for evaluation: the multi-hop QA dataset Bamboogle \cite{press-etal-2023-measuring}, and three single-hop QA datasets, NQ \cite{kwiatkowski2019natural}, TriviaQA \cite{joshi2017triviaqa}, and PopQA \cite{mallen-etal-2023-trust}. All OOD evaluations use the checkpoint trained on HotpotQA.

\textbf{Backbones.} We use Qwen2.5-7B-Instruct \cite{qwen2025qwen25technicalreport} and Qwen3-8B \cite{yang2025qwen3technicalreport} with consistent hyperparameters. 
\textbf{VAKE-P} denotes the Priming-only checkpoint, and \textbf{VAKE} denotes the checkpoint after both Priming and Reasoning. 
At test-time, VAKE uses either direct or insert-then-answer inference, while VAKE-P uses insert-then-answer inference with a frozen untrained answerer.

\textbf{Baselines.} All baselines take the question and the same retrieved subgraph, without extra retrieval, in three families. Direct answering: \textit{Base} answers directly and \textit{Base with Reasoning} adds CoT, prompted for Qwen2.5-7B and native for Qwen3-8B. Inference-time methods: \textit{Self-Ask} \citep{press-etal-2023-measuring} and \textit{RECITE} \citep{sun2023recitationaugmented}. Training-time (RL) methods: \textit{Unlock} \citep{yang2026reasoningreinforcementlearningunlocks} applies GRPO without insertion using a 72B judge, and \textit{GRPO} is a standard baseline for Qwen2.5-7B under the same RL budget, evaluated with direct answering or CoT.

\textbf{Graph construction and retrieval.}
We build $\mathcal{G}$ per dataset with an LLM-based OpenIE pipeline that extracts entities from each passage and then extracts triples conditioned on those entities. For a question $q$, we anchor its mentions to nodes in $\mathcal{G}$ by string matching and expand a two-hop BFS neighborhood keeping the top five edges per hop by cosine similarity to $q$, without reranking or query rewriting. We then drop any triple containing the gold answer and cap the remaining union at $K=10$ triples to form $\mathcal{R}(q)$, leaving it insufficient by construction.

\textbf{Evaluation metrics.}
Unless otherwise stated, tables report \textit{judge-EM}: given $q$, the
prediction $\hat a$, the reference $a^\star$, and any dataset-provided aliases,
a GPT-4o judge (temperature 0) outputs a binary semantic-equivalence decision,
accepting surface variants but
rejecting entity substitutions or factual drift. The judge sees neither the
retrieved subgraph nor the inserted triples, decoupling scoring from the
intervention. The prompt and rubric are in the Supplementary Material.

\textbf{Implementation Details.}
For each in-distribution dataset, we train VAKE with about 14K training examples and 200 validation examples. Each run uses one epoch, a batch size of 28, and a learning rate of $1\times10^{-6}$, yielding roughly 500 training steps. The policy samples 8 priming trajectories per step at temperature 0.7, with maximum prompt and response lengths of 2048 tokens. The answer-quality reward weights exact match and token-level F1 by 0.6 and 0.4, respectively; malformed outputs receive a $-1.0$ format penalty.

\subsection{Main Results}
\label{sec:main}
We compare VAKE with inference-time and training-time baselines on in-distribution and OOD QA benchmarks in Table~\ref{tab:main_results}, using matched test-time pipelines where applicable.

\textbf{Priming provides consistent gains} 
With Priming alone and no reasoning at inference, VAKE-P attains the best in-distribution average on both backbones, exceeding the strongest non-reasoning baseline by 2.4 and 2.2 points on Qwen2.5-7B and Qwen3-8B. 
The largest dataset-level margins appear on HotpotQA, where VAKE-P exceeds the strongest non-reasoning baseline by 2.3 and 2.2 points on the two backbones, respectively.
Since VAKE-P keeps the answerer frozen and changes only its input, the gains reflect the inserted bridging triples, not any adaptation of the answerer. Whether those triples supply retrievable facts or knowledge held in the model's parameters is the question we take up in \S\ref{sec:attribution}.

\textbf{Priming and Reasoning provide complementary gains.}
Under bridging-triple insertion with CoT at inference, VAKE outperforms VAKE-P across all in-distribution datasets and both backbones, raising the average by 1.9 points on Qwen2.5-7B and 1.7 on Qwen3-8B. This gain isolates the contribution of the Reasoning stage under a fixed test-time pipeline.  In the contrasting CoT-only setting, with no triples inserted, VAKE surpasses GRPO, the strongest reasoning baseline, by 6.1 and 2.6 points on the two backbones. Because this gain arises without any inserted triples, it reflects reasoning improved by Priming rather than extra evidence supplied at inference. We test this complementarity more directly through controlled ablations in \S\ref{sec:ablation}.

\textbf{VAKE generalizes across OOD tasks.}
VAKE achieves the best OOD average on both backbones, reaching 38.9 on Qwen2.5-7B and 41.4 on Qwen3-8B, and surpasses the strongest inference-time and RL baselines. 
The advantage is clearest on Qwen3-8B, where VAKE improves over the best competing method by 1.8 points on average and ranks first on all four OOD datasets. 
The largest gain appears on Bamboogle, the held-out multi-hop benchmark, where VAKE outperforms the strongest baseline by 2.4 points. This result suggests that the learned activation capability transfers beyond the training distribution and remains effective when accessing parametric knowledge across longer reasoning chains.

\subsection{Robustness: Scaling and General Capabilities}
\label{sec:exp_robustness}

This section studies whether VAKE remains stable across backbone sizes without sacrificing broader model capabilities.
For the scaling and input-condition analyses, we evaluate the VAKE-P checkpoint equipped with bridging-triple insertion. We further permit triple generation to be conditioned on raw documents, while the corresponding frozen base answerer still receives only the augmented subgraph.
\begin{table}[b]
\centering
\small
\renewcommand{\arraystretch}{1.15}
\begin{tabular*}{0.8\textwidth}{@{\extracolsep{\fill}} lcccccc}
\toprule
\textbf{Method}
& \textbf{GSM8K}
& \textbf{IFEval}
& \textbf{AIME24}
& \textbf{AIME25}
& \textbf{MMLU}
& \textbf{AVG} \\
\midrule

Base   & \underline{92.04} & 75.99 & \underline{33.33} & 23.33 & 70.84 & 59.11 \\
Unlock & 91.28 & 75.20 & 30.00 & 23.33 & 70.84 & 58.13 \\
GRPO   & \textbf{92.42} & \underline{76.52} & 23.33 & \textbf{30.00} & 70.62 & 58.58 \\
VAKE-P & \textbf{92.42} & 75.64 & \textbf{36.67} & 20.00 & \textbf{70.99} & \underline{59.14} \\
VAKE   & 91.74 & \textbf{76.82} & 30.00 & \underline{26.67} & \underline{70.87} & \textbf{59.22} \\

\bottomrule
\end{tabular*}
\caption{
    \textbf{General capability evaluation on standard benchmarks.}
    All scores are reported as percentages using each benchmark's standard
    evaluation metric. The \textbf{best} and \underline{second-best} results in
    each column are highlighted.
}
\label{tab:general_benchmarks}
\end{table}
\begin{figure}[h]
\centering
\includegraphics[width=0.7\columnwidth]{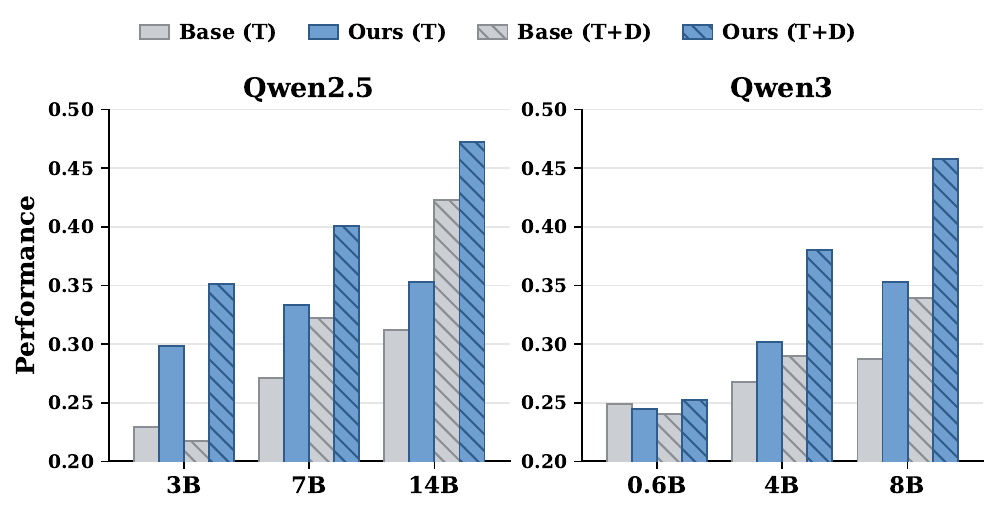}
\caption{Judge-EM of VAKE-P with bridging-triple insertion on 2Wiki across six Qwen backbones, using triples only (T) or triples with raw documents (T+D).}

\label{fig3:scaling}
\end{figure}

\textbf{Priming gains persist across backbone families and sizes.}

Figure~\ref{fig3:scaling} reports judge-EM on 2Wiki across six Qwen backbones under triples-only and document-augmented inputs for bridging-triple insertion. In the document-augmented setting, raw documents are available only to VAKE-P during triple generation; the corresponding frozen base answerer still predicts from the subgraph after triple insertion. VAKE-P outperforms Base in nearly all settings, except for the smallest Qwen3-0.6B model, where limited capacity and longer inputs may hurt triple-generation quality. The gains remain visible as model size increases, showing that stronger backbones still benefit from bridging-triple insertion. Adding raw documents further improves VAKE-P in several settings without exposing documents to the answerer, suggesting that extra context helps generate more effective bridging triples.

\textbf{VAKE preserves general capabilities.}
Table~\ref{tab:general_benchmarks} spans three categories: math, instruction following, and general knowledge. We evaluate on GSM8K \citep{cobbe2021training}, AIME24/25 \citep{aime24,aime25}, IFEval \citep{zhou2023instruction}, and MMLU \citep{hendrycks2020measuring}, reporting pass@10 on AIME24/25 and accuracy on the rest. VAKE-P and VAKE stay on par with Base across all three categories, with near-identical average scores, showing that the QA gains come at no cost to the evaluated general capabilities.

\subsection{Attribution of the Activated Knowledge}
\label{sec:attribution}
We next ask whether the inserted triples restate the retrieved subgraph or expose parametric knowledge.All analyses in this section are conducted on 2Wiki.



\textbf{Inserted triples mostly originate from parametric knowledge.}
For each backbone scale, we run LLM-based source attribution on the triples inserted by its own VAKE-P checkpoint. Given the retrieved subgraph and an inserted triple, the same GPT-4o judge labels the triple as \textit{subgraph-inferable} if it can be derived from the retrieved subgraph, and as \textit{parameter-originated} otherwise; the judge prompt is in the Supplementary Material. Across all scales, over 80\% of inserted triples are judged parameter-originated, and fewer than 20\% are attributed to the retrieved subgraph, as shown in Table~\ref{tab:insert-source}. Thus, most insertions cannot be explained as restatements or derivations of the retrieved subgraph and are operationally attributed to parametric knowledge.

\begin{table}[h]
\centering
\small
\renewcommand{\arraystretch}{1.2}
\begin{tabular*}{0.7\textwidth}{@{\extracolsep{\fill}} c cccccc}
\toprule
& \multicolumn{3}{c}{\textbf{Qwen2.5}} & \multicolumn{3}{c}{\textbf{Qwen3}} \\
\cmidrule(lr){2-4} \cmidrule(lr){5-7}
\textbf{Source} & \textbf{3B} & \textbf{7B} & \textbf{14B} & \textbf{0.6B} & \textbf{4B} & \textbf{8B} \\
\midrule
Params & 84.2 & 90.8 & 92.3 & 95.0 & 95.8 & 80.3 \\
Subgraph & 15.8 &  9.2 &  7.7 &  5.0 &  4.2 & 19.7 \\
\bottomrule
\end{tabular*}
\caption{Source attribution of \textsc{VAKE-P}-inserted triples across Qwen2.5/Qwen3 scales. Each cell reports the percentage of inserted triples attributed to model parameters or retrieved context.}
\label{tab:insert-source}
\end{table}

\begin{figure}[t]
\centering
\includegraphics[width=0.7\columnwidth]{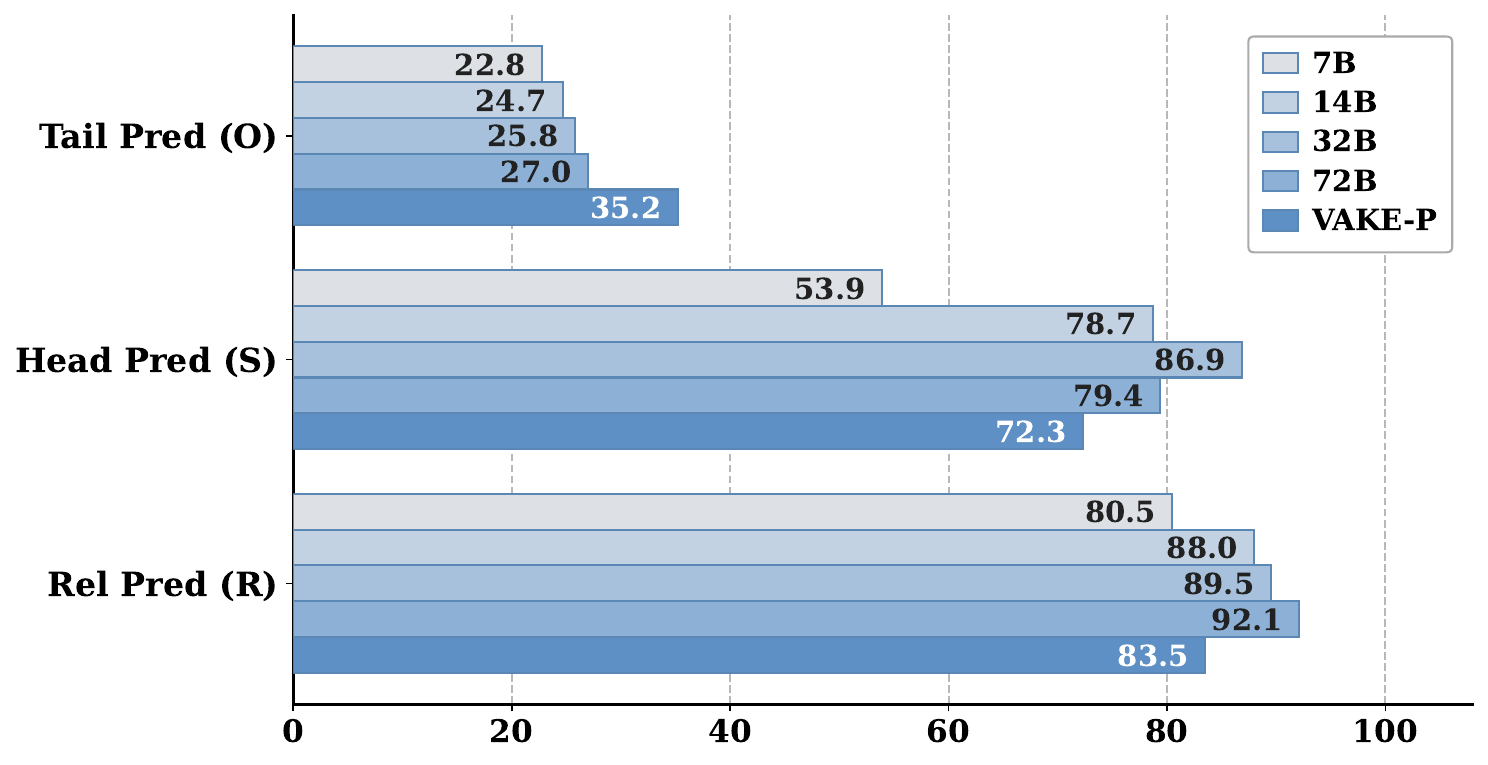}
\caption{Probing masked-slot reproduction of VAKE-P triples on 2Wiki across VAKE-P(7B) and frozen backbones.}

\label{fig:probe}
\end{figure}

\textbf{Probing reveals a gap between knowledge encoding and access.}
For the factually correct, parameter-originated triples inserted by VAKE-P (7B), we mask the head (S), relation (R), or tail (O) of each triple $(s,r,o)$ and measure \textit{reproduction accuracy}. We compare VAKE-P with frozen same-family backbones from 7B to 72B (Figure~\ref{fig:probe}). Frozen backbones achieve high head and relation accuracy, indicating that much of the relevant knowledge is encoded in their parameters. However, their tail accuracy stays below 27.0, whereas VAKE-P reaches 35.2 and outperforms even the 72B backbone. This contrast suggests that Priming improves access to encoded answer-bearing information.

\subsection{Ablation Studies}
\label{sec:ablation}
We ablate VAKE on Qwen2.5-7B-Instruct along two axes: its training components and input conditions.

\textbf{Priming and Reasoning training are complementary.}

Table~\ref{tab:method-ablation} ablates VAKE's two training stages. All variants are evaluated without bridging-triple insertion; VAKE and \textit{w/o Priming} use CoT reasoning. VAKE performs best on all datasets, while \textit{w/o Reasoning} surpasses both Base and \textit{w/o Priming} on HotpotQA and MuSiQue, showing that Priming benefits direct answering even without triple insertion.
\begin{table}[h]
\centering
\small
\renewcommand{\arraystretch}{1.1}
\begin{tabular*}{0.7\textwidth}{@{\extracolsep{\fill}} lccc}
\toprule
\textbf{Variant} & \textbf{2Wiki} & \textbf{HotpotQA} & \textbf{MuSiQue} \\
\midrule
VAKE
    & \textbf{35.50}
    & \textbf{43.40}
    & \textbf{20.40} \\
\quad w/o Priming
    & \underline{33.50}
    & 31.90
    & 15.80 \\
\quad w/o Reasoning
    & 28.50
    & \underline{37.20}
    & \underline{16.40} \\
\quad Base
    & 27.20
    & 32.70
    & 13.40 \\
\bottomrule
\end{tabular*}
\caption{Method ablation on multi-hop QA. \textit{w/o Priming} and \textit{w/o Reasoning} remove the corresponding training stage. All variants are evaluated without bridging-triple insertion.}
\label{tab:method-ablation}
\end{table}

\begin{table}[t]
\centering
\small
\renewcommand{\arraystretch}{1.1}
\begin{tabular*}{0.7\textwidth}{@{\extracolsep{\fill}} l c c c}
\toprule
\textbf{Method}
 & \textbf{Closed-book} & \textbf{Triples} & \textbf{Triples+Docs} \\
\midrule
Base            & 24.20           & 27.60           & 30.70           \\
VAKE-P          & 30.00           & 33.10           & 41.30           \\
VAKE    & \textbf{30.10} & \textbf{36.20} & \textbf{46.60} \\
\bottomrule
\end{tabular*}
\caption{Information ablation on 2Wiki, with all methods evaluated via bridging-triple insertion.}
\label{tab:info-ablation}
\end{table}

\textbf{The learned activation capability generalizes across input settings.}

Table~\ref{tab:info-ablation} compares Base, VAKE-P, and VAKE on 2Wiki across closed-book, triples-only, and document-augmented inputs. All methods use bridging-triple insertion, thereby isolating the effect of the trained checkpoints under the same test-time pipeline. VAKE-P and VAKE improve over Base in all three settings, confirming that the gains arise from training rather than triple insertion alone. The closed-book improvement further supports parametric knowledge elicitation, as bridging triples are generated from the question without retrieved context. Under document-augmented input, VAKE improves over Base by 15.9 points, showing that the learned activation capability also complements richer external context.


\section{Conclusion}

In this paper, we frame factual question answering as a problem of accessing parametric knowledge rather than storing it, as encoded facts often remain implicit and inaccessible. We propose VAKE, a two-stage reinforcement-learning framework that separates explicit knowledge elicitation from direct answer reasoning. In Priming, the policy inserts bridging triples into a sparse retrieved subgraph while a frozen answerer provides outcome-based rewards, rendering the activated knowledge observable and its answer-enabling effect directly testable. In Reasoning, the elicitation capability learned during Priming transfers to direct chain-of-thought answering and combines with reasoning-oriented optimization. Experiments on two instruction-tuned backbones across seven multi-hop and single-hop benchmarks show that VAKE consistently outperforms inference-time and reinforcement-learning baselines and generalizes to OOD datasets. 
A scaling analysis across six Qwen backbones shows consistent activation gains at larger scales.
Attribution analysis indicates that the inserted triples predominantly surface parametric knowledge inaccessible through direct prompting rather than restating retrieved context.
VAKE can be integrated into existing post-training pipelines as an RL-based complement to QA-pair construction and SFT-based knowledge activation. Our ablations further show that Priming improves direct answering without test-time triple insertion, while the learned activation capability generalizes from closed-book to document-augmented settings.

\clearpage

\bibliographystyle{plainnat}
\bibliography{main}

@article{calderon2026empty,
  title={Empty shelves or lost keys? recall is the bottleneck for parametric factuality},
  author={Calderon, Nitay and Ben-David, Eyal and Gekhman, Zorik and Ofek, Eran and Yona, Gal},
  journal={arXiv preprint arXiv:2602.14080},
  year={2026}
}

@inproceedings{wang-etal-2024-unveiling,
    title={Unveiling Factual Recall Behaviors of Large Language Models through Knowledge Neurons},
    author={Wang, Yifei and Chen, Yuheng and Wen, Wanting and Sheng, Yu and Li, Linjing and Zeng, Daniel Dajun},
    booktitle={Proceedings of the Conference on Empirical Methods in Natural Language Processing (EMNLP)},
    pages={7388--7402},
    year={2024}
}

@inproceedings{orgad2025llms,
    title={{LLM}s Know More Than They Show: On the Intrinsic Representation of {LLM} Hallucinations},
    author={Hadas Orgad and Michael Toker and Zorik Gekhman and Roi Reichart and Idan Szpektor and Hadas Kotek and Yonatan Belinkov},
    booktitle={International Conference on Learning Representations (ICLR)},
    year={2025},
}

@inproceedings{gekhman2025insideout,
    title={Inside-Out: Hidden Factual Knowledge in {LLM}s},
    author={Zorik Gekhman and Eyal Ben-David and Hadas Orgad and Eran Ofek and Yonatan Belinkov and Idan Szpektor and Jonathan Herzig and Roi Reichart},
    booktitle={Conference on Language Modeling (COLM)},
    year={2025},
}

@inproceedings{liu-etal-2025-structure,
  title={Structure-aware domain knowledge injection for large language models},
  author={Liu, Kai and Chen, Ze and Fu, Zhihang and Zhang, Wei and Jiang, Rongxin and Zhou, Fan and Chen, Yaowu and Wu, Yue and Ye, Jieping},
  booktitle={Proceedings of the Annual Meeting of the Association for Computational Linguistics (ACL)"},
  pages={29443--29464},
  year={2025}
}

@inproceedings{sun2023recitationaugmented,
    title={Recitation-Augmented Language Models},
    author={Zhiqing Sun and Xuezhi Wang and Yi Tay and Yiming Yang and Denny Zhou},
    booktitle={International Conference on Learning Representations (ICLR)},
    year={2023},
}

@inproceedings{liu-etal-2022-generated,
    title = "Generated Knowledge Prompting for Commonsense Reasoning",
    author = "Liu, Jiacheng  and Liu, Alisa  and Lu, Ximing  and Welleck, Sean  and
      West, Peter  and
      Le Bras, Ronan  and
      Choi, Yejin  and
      Hajishirzi, Hannaneh",
    booktitle = "Proceedings of the Annual Meeting of the Association for Computational Linguistics (ACL)",
    year = "2022",
}

@inproceedings{NEURIPS2022_9d560961,
    author = {Wei, Jason and Wang, Xuezhi and Schuurmans, Dale and Bosma, Maarten and Ichter, Brian and Xia, Fei and Chi, Ed and Le, Quoc V and Zhou, Denny},
    booktitle = {Advances in Neural Information Processing Systems (NeurIPS)},
    pages = {24824--24837},
    title = {Chain-of-Thought Prompting Elicits Reasoning in Large Language Models},
    volume = {35},
    year = {2022}
}

@inproceedings{zhong-etal-2021-factual,
    title={Factual probing is [mask]: Learning vs. learning to recall},
    author={Zhong, Zexuan and Friedman, Dan and Chen, Danqi},
    booktitle={Proceedings of the Conference of the North American Chapter of the Association for Computational Linguistics: Human Language Technologies (NAACL-HLT)},
    pages={5017--5033},
    year={2021}
}

@inproceedings{shin-etal-2020-autoprompt,
    title={Autoprompt: Eliciting knowledge from language models with automatically generated prompts},
    author={Shin, Taylor and Razeghi, Yasaman and Iv, Robert L Logan and Wallace, Eric and Singh, Sameer},
    booktitle={Proceedings of the Conference on Empirical Methods in Natural Language Processing (EMNLP)},
    pages={4222--4235},
    year={2020}
}

@inproceedings{azaria-mitchell-2023-internal,
    title={The internal state of an LLM knows when it’s lying},
    author={Azaria, Amos and Mitchell, Tom},
    booktitle={Proceedings of the Conference on Empirical Methods in Natural Language Processing (EMNLP)},
    pages={967--976},
    year={2023}
}

@article{burns2023discovering,
    title={Discovering latent knowledge in language models without supervision},
    author={Burns, Collin and Ye, Haotian and Klein, Dan and Steinhardt, Jacob},
    journal={arXiv preprint arXiv:2212.03827},
    year={2022}
}

@inproceedings{gekhman-etal-2024-fine,
    title={Does fine-tuning llms on new knowledge encourage hallucinations?},
    author={Gekhman, Zorik and Yona, Gal and Aharoni, Roee and Eyal, Matan and Feder, Amir and Reichart, Roi and Herzig, Jonathan},
    booktitle={Proceedings of the Conference on Empirical Methods in Natural Language Processing (EMNLP)},
    pages={7765--7784},
    year={2024}
}

@article{elazar-etal-2021-measuring,
    title = "Measuring and Improving Consistency in Pretrained Language Models",
    author = {Elazar, Yanai  and
      Kassner, Nora  and
      Ravfogel, Shauli  and
      Ravichander, Abhilasha  and
      Hovy, Eduard  and
      Sch{\"u}tze, Hinrich  and
      Goldberg, Yoav},
    journal = "Transactions of the Association for Computational Linguistics (TACL)",
    volume = "9",
    year = "2021",
    pages = "1012--1031",
}

@article{jiang-etal-2020-know,
    title = "How Can We Know What Language Models Know?",
    author = "Jiang, Zhengbao  and
      Xu, Frank F.  and
      Araki, Jun  and
      Neubig, Graham",
    journal = "Transactions of the Association for Computational Linguistics (TACL)",
    volume = "8",
    year = "2020",
    pages = "423--438",
}

@inproceedings{zheng2024take,
    title={Take a Step Back: Evoking Reasoning via Abstraction in Large Language Models},
    author={Huaixiu Steven Zheng and Swaroop Mishra and Xinyun Chen and Heng-Tze Cheng and Ed H. Chi and Quoc V Le and Denny Zhou},
    booktitle={International Conference on Learning Representations (ICLR)},
    year={2024},
}

@inproceedings{NEURIPS2023_81b83900,
    author = {Li, Kenneth and Patel, Oam and Vi\'{e}gas, Fernanda and Pfister, Hanspeter and Wattenberg, Martin},
    booktitle = {Advances in Neural Information Processing Systems (NeurIPS)},
    pages = {41451--41530},
    title = {Inference-Time Intervention: Eliciting Truthful Answers from a Language Model},
    volume = {36},
    year = {2023}
}

@inproceedings{
    li2025reasoning,
    title={Reasoning Models Hallucinate More: Factuality-Aware Reinforcement Learning for Large Reasoning Models},
    author={Junyi Li and Hwee Tou Ng},
    booktitle={Advances in Neural Information Processing Systems (NeurIPS)},
    year={2025},
    url={https://openreview.net/forum?id=Igq7Dyc3OL}
}

@inproceedings{ren-etal-2026-knowrl,
    title = "{K}now{RL}: Exploring Knowledgeable Reinforcement Learning for Factuality",
    author = "Ren, Baochang  and
      Qiao, Shuofei  and
      Zhang, Ningyu  and
      Zheng, Da  and
      Chen, Huajun",
    booktitle = "Proceedings of the Annual Meeting of the Association for Computational Linguistics (ACL)",
    year = "2026",
    pages = "39640--39658"
}

@inproceedings{
    wen2026reinforcement,
    title={Reinforcement Learning with Verifiable Rewards Implicitly Incentivizes Correct Reasoning in Base {LLM}s},
    author={Xumeng Wen and Zihan Liu and Shun Zheng and Shengyu Ye and Zhirong Wu and Yang Wang and Zhijian Xu and Xiao Liang and Junjie Li and Ziming Miao and Jiang Bian and Mao Yang},
    booktitle={International Conference on Learning Representations (ICLR)},
    year={2026},
    url={https://openreview.net/forum?id=jGbRWwIidy}
}

@inproceedings{
    yue2025does,
    title={Does Reinforcement Learning Really Incentivize Reasoning Capacity in {LLM}s Beyond the Base Model?},
    author={Yang Yue and Zhiqi Chen and Rui Lu and Andrew Zhao and Zhaokai Wang and Yang Yue and Shiji Song and Gao Huang},
    booktitle={Advances in Neural Information Processing Systems (NeurIPS)},
    year={2025},
    url={https://openreview.net/forum?id=4OsgYD7em5}
}

@inproceedings{wang2023selfconsistency,
    title={Self-Consistency Improves Chain of Thought Reasoning in Language Models},
    author={Xuezhi Wang and Jason Wei and Dale Schuurmans and Quoc V Le and Ed H. Chi and Sharan Narang and Aakanksha Chowdhery and Denny Zhou},
    booktitle={International Conference on Learning Representations (ICLR)},
    year={2023},
    url={https://openreview.net/forum?id=1PL1NIMMrw}
}

@article{wei2025truthrl,
  title={Truthrl: Incentivizing truthful llms via reinforcement learning},
  author={Wei, Zhepei and Yang, Xiao and Sun, Kai and Wang, Jiaqi and Shao, Rulin and Chen, Jingxiang and Kachuee, Mohammad and Gollapudi, Teja and Liao, Yiwei and Scheffer, Nicolas and others},
  journal={arXiv preprint arXiv:2509.25760},
  year={2025}
}

@article{chen2025learning,
  title={Learning to reason for factuality},
  author={Chen, Xilun and Kulikov, Ilia and Berges, Vincent-Pierre and O{\u{g}}uz, Barlas and Shao, Rulin and Ghosh, Gargi and Weston, Jason and Yih, Wen-tau},
  journal={arXiv preprint arXiv:2508.05618},
  year={2025}
}

@article{ma2026improving,
  title={Improving parametric knowledge access in reasoning language models},
  author={Ma, Melody and Hewitt, John},
  journal={arXiv preprint arXiv:2602.22193},
  year={2026}
}

@article{gekhman2026thinking,
  title={Thinking to recall: How reasoning unlocks parametric knowledge in llms},
  author={Gekhman, Zorik and Aharoni, Roee and Ofek, Eran and Geva, Mor and Reichart, Roi and Herzig, Jonathan},
  journal={arXiv preprint arXiv:2603.09906},
  year={2026}
}

@article{ovadia2025knowledge,
  title={Knowledge-instruct: Effective continual pre-training from limited data using instructions},
  author={Ovadia, Oded and Brief, Meni and Lemberg, Rachel and Sheetrit, Eitam},
  journal={arXiv preprint arXiv:2504.05571},
  year={2025}
}

@article{meng2026sparse,
  title={Sparse but critical: A token-level analysis of distributional shifts in rlvr fine-tuning of llms},
  author={Meng, Haoming and Huang, Kexin and Wei, Shaohang and Ma, Chiyu and Yang, Shuo and Wang, Xue and Wang, Guoyin and Ding, Bolin and Zhou, Jingren},
  journal={arXiv preprint arXiv:2603.22446},
  year={2026}
}

@misc{yang2026reasoningreinforcementlearningunlocks,
      title={Beyond Reasoning: Reinforcement Learning Unlocks Parametric Knowledge in LLMs}, 
      author={Wanli Yang and Hongyu Zang and Junwei Zhang and Wenjie Shi and Du Su and Jingang Wang and Xueqi Cheng and Fei Sun},
      year={2026},
      eprint={2605.07153},
      archivePrefix={arXiv},
      primaryClass={cs.CL},
      note={arXiv preprint arXiv:2605.07153},
      url={https://arxiv.org/abs/2605.07153},
}

@article{mousavi2026doeslossoptimizationactually,
  title={What Does Loss Optimization Actually Teach, If Anything? Knowledge Dynamics in Continual Pre-training of LLMs},
  author={Mousavi, Seyed Mahed and Alghisi, Simone and Riccardi, Giuseppe},
  journal={arXiv preprint arXiv:2601.03858},
  year={2026}
}

@article{zheng2024reliablellmsknowledgebases,
  title={How reliable are LLMs as knowledge bases? Re-thinking facutality and consistency},
  author={Zheng, Danna and Lapata, Mirella and Pan, Jeff Z},
  journal={arXiv preprint arXiv:2407.13578},
  year={2024}
}

@article{deepseekai2026deepseekv4highlyefficientmilliontoken,
  title={DeepSeek-v4: Towards highly efficient million-token context intelligence},
  author={Xu, Anyi and Lin, Bangcai and Xue, Bing and Wang, Bingxuan and Xu, Bingzheng and Wu, Bochao and Zhang, Bowei and Lin, Chaofan and Dong, Chen and Ling, Chenchen and others},
  journal={arXiv preprint arXiv:2606.19348},
  year={2026}
}

@article{glm5team2026glm5vibecodingagentic,
  title={GLM-5: from vibe coding to agentic engineering},
  author={Zeng, Aohan and Lv, Xin and Hou, Zhenyu and Du, Zhengxiao and Zheng, Qinkai and Chen, Bin and Yin, Da and Ge, Chendi and Huang, Chenghua and Xie, Chengxing and others},
  journal={arXiv preprint arXiv:2602.15763},
  year={2026}
}

@article{yang2025qwen3technicalreport,
  title={Qwen3 technical report},
  author={Yang, An and Li, Anfeng and Yang, Baosong and Zhang, Beichen and Hui, Binyuan and Zheng, Bo and Yu, Bowen and Gao, Chang and Huang, Chengen and Lv, Chenxu and others},
  journal={arXiv preprint arXiv:2505.09388},
  year={2025}
}

@inproceedings{press-etal-2023-measuring,
    title = "Measuring and Narrowing the Compositionality Gap in Language Models",
    author = "Press, Ofir  and
      Zhang, Muru  and
      Min, Sewon  and
      Schmidt, Ludwig  and
      Smith, Noah  and
      Lewis, Mike",
    booktitle = "Proceedings of the Conference on Empirical Methods in Natural Language Processing (EMNLP)",
    year = "2023",
    pages = "5687--5711",
}

@inproceedings{ho2020constructing,
  title={Constructing a multi-hop qa dataset for comprehensive evaluation of reasoning steps},
  author={Ho, Xanh and Nguyen, Anh-Khoa Duong and Sugawara, Saku and Aizawa, Akiko},
  booktitle={Proceedings of the 28th International Conference on Computational Linguistics},
  pages={6609--6625},
  year={2020}
}

@inproceedings{yang2018hotpotqa,
  title={HotpotQA: A dataset for diverse, explainable multi-hop question answering},
  author={Yang, Zhilin and Qi, Peng and Zhang, Saizheng and Bengio, Yoshua and Cohen, William and Salakhutdinov, Ruslan and Manning, Christopher D},
  booktitle={Proceedings of the Conference on Empirical Methods in Natural Language Processing (EMNLP)},
  pages={2369--2380},
  year={2018}
}

@article{trivedi2022musique,
  title={MuSiQue: Multihop Questions via Single-hop Question Composition},
  author={Trivedi, Harsh and Balasubramanian, Niranjan and Khot, Tushar and Sabharwal, Ashish},
  journal={Transactions of the Association for Computational Linguistics},
  volume={10},
  pages={539--554},
  year={2022},
  publisher={MIT Press One Broadway, 12th Floor, Cambridge, Massachusetts 02142, USA~…}
}

@article{kwiatkowski2019natural,
  title={Natural questions: a benchmark for question answering research},
  author={Kwiatkowski, Tom and Palomaki, Jennimaria and Redfield, Olivia and Collins, Michael and Parikh, Ankur and Alberti, Chris and Epstein, Danielle and Polosukhin, Illia and Devlin, Jacob and Lee, Kenton and others},
  journal={Transactions of the Association for Computat
  ional Linguistics},
  volume={7},
  pages={453--466},
  year={2019},
  publisher={MIT Press One Rogers Street, Cambridge, MA 02142-1209, USA journals-info~…}
}

@inproceedings{mallen-etal-2023-trust,
    title = "When Not to Trust Language Models: Investigating Effectiveness of Parametric and Non-Parametric Memories",
    author = "Mallen, Alex  and
      Asai, Akari  and
      Zhong, Victor  and
      Das, Rajarshi  and
      Khashabi, Daniel  and
      Hajishirzi, Hannaneh",
    booktitle = "Proceedings of the Annual Meeting of the Association for Computational Linguistics (ACL)",
    year = "2023",
    pages = "9802--9822",
}

@inproceedings{joshi2017triviaqa,
  title={Triviaqa: A large scale distantly supervised challenge dataset for reading comprehension},
  author={Joshi, Mandar and Choi, Eunsol and Weld, Daniel S and Zettlemoyer, Luke},
  booktitle={Proceedings of the 55th Annual Meeting of the Association for Computational Linguistics (Volume 1: Long Papers)},
  pages={1601--1611},
  year={2017}
}

@misc{qwen2025qwen25technicalreport,
      title={Qwen2.5 Technical Report}, 
      author={Qwen and An Yang and Baosong Yang and Beichen Zhang and Binyuan Hui and Bo Zheng and Bowen Yu and others},
      year={2025},
      eprint={2412.15115},
      archivePrefix={arXiv},
      primaryClass={cs.CL},
      url={https://arxiv.org/abs/2412.15115}, 
}

@article{cobbe2021training,
  title={Training verifiers to solve math word problems},
  author={Cobbe, Karl and Kosaraju, Vineet and Bavarian, Mohammad and Chen, Mark and Jun, Heewoo and Kaiser, Lukasz and Plappert, Matthias and Tworek, Jerry and Hilton, Jacob and Nakano, Reiichiro and others},
  journal={arXiv preprint arXiv:2110.14168},
  year={2021}
}

@article{zhou2023instruction,
  title={Instruction-following evaluation for large language models},
  author={Zhou, Jeffrey and Lu, Tianjian and Mishra, Swaroop and Brahma, Siddhartha and Basu, Sujoy and Luan, Yi and Zhou, Denny and Hou, Le},
  journal={arXiv preprint arXiv:2311.07911},
  year={2023}
}

@misc{aime24,
      title={American Invitational Mathematics Examination (AIME) 2024}, 
      author={Zhang, Yifan and Math-AI, Team},
      year={2024},
}

@misc{aime25,
      title={American Invitational Mathematics Examination (AIME) 2025}, 
      author={Zhang, Yifan and Math-AI, Team},
      year={2025},
}

@article{hendrycks2020measuring,
  title={Measuring massive multitask language understanding},
  author={Hendrycks, Dan and Burns, Collin and Basart, Steven and Zou, Andy and Mazeika, Mantas and Song, Dawn and Steinhardt, Jacob},
  journal={arXiv preprint arXiv:2009.03300},
  year={2020}
}

\appendix

\end{document}